%% file: main.tex
\documentclass[10pt,twocolumn,letterpaper]{article}

\usepackage[pagenumbers]{iccv} 
\usepackage{multirow}
\usepackage{tabularray}
\usepackage{bbding}
\usepackage{siunitx} 
\usepackage{enumitem}
\newcommand{\minisection}[1]{\vspace{0.04in} \noindent {\bf #1}\ \ }

\input{preamble}
\usepackage[noend]{algpseudocode}
\usepackage{algorithm}
\definecolor{iccvblue}{rgb}{0.21,0.49,0.74}
\usepackage[pagebackref,breaklinks,colorlinks,allcolors=iccvblue]{hyperref}

\def\paperID{12906} 
\def\confName{ICCV}
\def\confYear{2025}

\title{Knowledge Graph Enhanced Generative Multi-modal Models for Class-Incremental Learning}

\author{Xusheng Cao$^1$, Haori Lu$^1$, Linlan Huang$^1$, Fei Yang$^{2,1}$, Xialei Liu$^{2,1}$, Ming-Ming Cheng$^{2,1}$ \\
$^1$VCIP, CS, Nankai University \qquad $^2$NKIARI, Shenzhen Futian\\
{\tt\small \{caoxusheng, luhaori, huanglinlan\}@mail.nankai.edu.cn, \{cmm, xialei\}@nankai.edu.cn}
}

\begin{document}
\maketitle
\input{sec/0_abstract}    
\input{sec/1_intro}

\input{sec/2_related_work}

\input{sec/3_method}
\input{sec/4_exp.tex}

\section{Conclusions}
In this paper, we propose KG-GMM, a novel method to combine MLLMs and knowledge graphs to tackle the catastrophic forgetting problem in continual learning. Our method gradually builds a graph in the process of learning new classes, assigning each class with $r$ different relations to enhance discriminability between semantic similar classes. During inference, we use the generated relations to locate the specific class, combined with the predicted text, our KG-GMM can effectively preserve much of the LLM’s generalization ability while providing more accurate category predictions for given test images. Extensive experiments show that our method outperforms the state-of-the-art baselines for exemplar-free class incremental learning.


\clearpage
{
    \small
    \bibliographystyle{ieeenat_fullname}
    \bibliography{main}
}

\end{document}

%% file: sec/0_abstract.tex
\begin{abstract}
Continual learning in computer vision faces the critical challenge of catastrophic forgetting, where models struggle to retain prior knowledge while adapting to new tasks.  
Although recent studies have attempted to leverage the generalization capabilities of pre-trained models to mitigate overfitting on current tasks, models still tend to forget details of previously learned categories as tasks progress, leading to misclassification. To address these limitations, we introduce a novel Knowledge Graph Enhanced Generative Multi-modal model (KG-GMM) that builds an evolving knowledge graph throughout the learning process.
Our approach utilizes relationships within the knowledge graph to augment the class labels and assigns different relations to similar categories to enhance model differentiation. During testing, we propose a Knowledge Graph Augmented Inference method that locates specific categories by analyzing relationships within the generated text, thereby reducing the loss of detailed information about old classes when learning new knowledge and alleviating forgetting. 
Experiments demonstrate that our method effectively leverages relational information to help the model correct mispredictions, achieving state-of-the-art results in both conventional CIL and few-shot CIL settings, confirming the efficacy of knowledge graphs at preserving knowledge in the continual learning scenarios.
\end{abstract}

%% file: sec/1_intro.tex
\section{Introduction}

Continual learning without forgetting old knowledge~\cite{mccloskey1989catastrophic} has been thriving in the ever-changing world.
Traditional approaches in this area typically employ three categories~\cite{masana2022class} of methods to prevent forgetting: replay-based methods~\cite{li2017learning, rebuffi2017icarl,UCIR_2019_CVPR, wu2019large} that retain a portion of past data, architecture-based methods~\cite{mallya2018piggyback, serra2018overcoming,yan2021dynamically} that progressively expand the model, and regularization-based methods~\cite{Chaudhry_2018_ECCV,kirkpatrick2017overcoming,aljundi2018memory} that prevent excessive changes in model parameters. Recently, methods based on large-scale pre-trained models~\cite{jia2021scaling,radford2021learning} have attracted more attention due to their superior performance than the traditional train-from-scratch methods. For example, prompt-based methods~\cite{smith2023coda,l2p,wang2022dualprompt,wang2023hide} have been proposed that use a small number of parameters to learn “general” and “specific” knowledge without altering the pre-trained backbone. SLCA~\cite{zhang2023slca}  proposes to use a small learning rate for the backbone to preserve the generalizability and a larger learning rate for the classifier to accommodate new classes. Yet, these methods do not leverage textual information of the class labels or the relationships between the learned classes.

\begin{figure}[t]
    \centering
    \begin{subfigure}[t]{0.4\columnwidth}
        \centering
        \includegraphics[width=\linewidth]{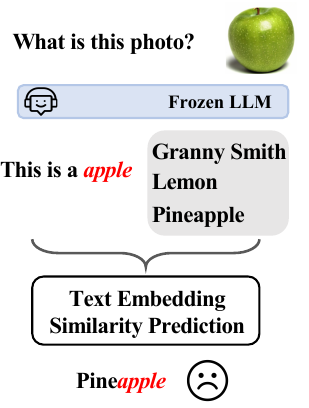}
        \caption{GMM inference pipeline.}
    \end{subfigure}%
    \hfill
    \begin{subfigure}[t]{0.55\columnwidth}
        \centering
        \includegraphics[width=\linewidth]{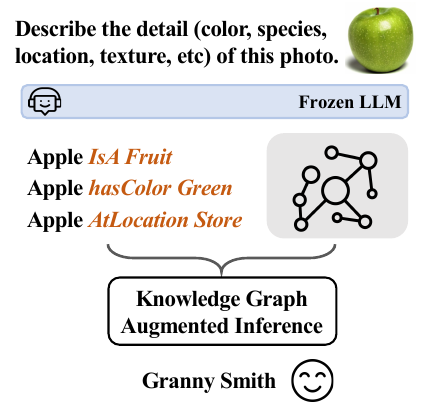}
        \caption{KG-GMM inference pipeline.}
    \end{subfigure}
    \caption{The difference of inference pipeline between the generative-based baseline GMM~\cite{cao2024GMM} and our Knowledge Graph enhanced generative multi-modal model (KG-GMM).}
    \label{fig:teaser}
    \vspace{-5mm}
\end{figure}

To leverage the rich information lies in text modality, Continual CLIP~\cite{thengane2022continualclip} proposes to use Frozen CLIP to conduct predictions based on image-text similarities. RAPF~\cite{huang2024rapf} uses CLIP~\cite{radford2021learning} text encoder to detect and pull apart similar classes. PROOF~\cite{zhou2023proof} proposes to use task-specific projections on both image encoder and text encoder. MoE-Adapters~\cite{yu2024boosting} utilize the Mixture-of-Experts (MoE) adapters on top of the pre-trained CLIP model and an Auto-Selector to route for different input.  CLIP-CIL~\cite{liu2023clipCIL} proposes to fine-tune an additional adapter after the backbone to learn new classes. However, in the more challenging exemplar-free continual learning scenario, these methods still face problems with classification bias towards the newly learned knowledge and forgetting of previously learned knowledge.
In contrast, GMM~\cite{cao2024GMM} addresses the bias problem by discarding the traditional classifier and directly using a Large Language Model (LLM) to generate predicted text. GMM employs image-text pairs as input and only fine-tunes a linear layer, effectively mitigating bias and leveraging the LLM’s capability to understand and generate human-interpretable text. However, two challenges persist during the learning process.

Firstly, fine-tuning a small linear layer with a single, fixed format could cause the Multi-modal LLM to lose its generalization capabilities, causing it to output text only in one format ``This is a photo of a [CLS]'' while losing the ability to describe details about the image content in terms of colors, background, texture, etc. Secondly, since the model lacks exposure to images from previous tasks, it tends to classify all images encountered into higher-level categories (knowledge obtained in the pre-training phase). For example, if the model learns ``Granny Smith'' in the initial task, it can accurately output ``Granny Smith'' for images of that class during immediate testing. 
However, after learning other similar categories like ``Pineapple'' or ``Lemon'' in subsequent tasks without replaying exemplars, the model tends to respond, ``This is a photo of a apple'' when presented with green apple images during testing and misclassifying it to a newly learned text related label ``Pineapple'' as shown in Fig.~\ref{fig:teaser}.

To address these challenges, we turn to common sense knowledge graphs~\cite{ilievski2021cskg}, which are structured collections of factual triples, organized as (head, relation, object), \eg, (\text{Granny Smith}, \text{IsA}, \text{Fruit}) that capture relationships between entities. Compared to the black-box generation process of LLMs~\cite{pan2024unifying}, knowledge graphs encode rich interrelationships information in the form of human-readable language. 
For example, ConceptNet~\cite{speer2017conceptnet} is a general common sense knowledge graph integrating triplets that cover nearly all visual recognition datasets (including both coarse-grained and fine-grained ones) with efficient query operations.
We believe that a knowledge graph that incrementally expands as tasks increase would be lightweight and straightforward while retaining the knowledge structures learned by the model to prevent forgetting.


Building on this idea, we propose a Knowledge Graph Enhanced Generative Multi-modal Model for class-incremental learning. It enables the model to focus more on the factual content within the images, providing descriptive output rather than direct guesses. Additionally, during inference, we construct a subgraph from the model’s original textual output and compare it with the existing graph to identify the specific category to which the image belongs.
Specifically, we use a common-sense knowledge graph to incrementally build a sub-graph during continual learning by storing class-relevant triplets. Training employs relations (not plain text) as ground truth labels. During testing, associative keywords (e.g., IsA, AtLocation) guide the model to output detailed image facts instead of direct guesses, enhancing precision by recalling prior relational knowledge from the graph. This enables structured retention and retrieval of learned relationships, improving answer specificity.

The main contributions of this paper are:

\begin{itemize}
    \item We propose using an ever-expanding knowledge graph to help the model distinguish similar classes across different tasks based on relationships, providing more references when discriminating similar classes.
    \item During inference, we propose to guide the model with relation keywords to output more factual descriptions rather than direct guesses, preventing forgetting through associative relationships.
    \item Experiments on multiple datasets and settings demonstrate that our method can help the model alleviate forgetting with minimal training cost.

\end{itemize}

%% file: sec/2_related_work.tex
\section{Related Work}

\subsection{Class Incremental Learning}
Class incremental learning is a method where a model learns new classes sequentially over time without forgetting the previously learned classes.
Early methods of class-incremental learning often involved training from scratch~\cite{kirkpatrick2017overcoming,douillard2021dytox,wang2022foster,zhu2021prototype,2024LDC,yu2020semantic}. 

With the advent of pre-trained models, using them as a starting point for continual learning has shown great effectiveness.
Initially, models pre-trained on ImageNet are typically fine-tuned using parameter-efficient methods to incorporate new knowledge, such as prompt-based techniques~\cite{2024convprompt,wang2022dualprompt,l2p,smith2023coda} and low-rank adapters~\cite{2024beyond,liang2024inflora}. 
Some approaches involve fine-tuning specific modules of the model~\cite{zhou2024revisiting} or setting differential learning rates~\cite{zhang2023slca}. 
Building on this, guidance from textual modality information further alleviates forgetting. The CLIP model, with its strong zero-shot capabilities, has significantly benefited class-incremental learning~\cite{thengane2022continualclip} and garnered considerable attention.
This is often achieved by adding parameter modules, such as linear layers~\cite{liu2023clipCIL,huang2024rapf} or attention modules~\cite{jha2024clap4clip,zhou2023proof}, to the original model to directly adjust the well-learned features.

Currently, generative multi-modal models demonstrate powerful capabilities at mitigating the forgetting problem in continual learning. 
Simple fine-tuning on these models can effectively retain old knowledge while learning new classes~\cite{cao2024GMM}. However, without data replay, the GMM still tends to forget details of the classes learned in former tasks. Our method tackles this issue by incorporating rich relationship information in knowledge graphs to enhance the GMM learning and inference process.

\subsection{Knowledge Graph}

Knowledge graphs are structured representations of information where entities (nodes) are interconnected through relationships (edges), allowing for complex querying and inference over linked data. There are four main types of knowledge graphs~\cite{pan2024unifying}: Encyclopedic KGs that cover general knowledge (\eg, Wikidata~\cite{vrandevcic2014wikidata}), Common Sense KGs (\eg, ConceptNet~\cite{speer2017conceptnet}) that capture everyday concepts and objects, Domain-specific KGs tailored to specialized fields (\eg, UMLS~\cite{bodenreider2004unified} for medical domain), and Multi-modal KGs ~\cite{liu2019mmkg} that integrate various data types such as text and images. We mainly focus on the Common Sense KG due to its broad scope that covers most classes in image datasets.

With the development of LLM, there has been research work that combines the generation capability of LLM models with the structured, rich factual knowledge stored in knowledge graphs. The combinations mainly fall into three categories: KG-enhanced LLMs~\cite{lin-etal-2019-kagnet,petroni2019language,zhang2024graphtranslator} involve embedding KGs to improve LLMs by enhancing understanding and reasoning of the knowledge learned by LLMs. LLM-augmented KGs~\cite{cao2022relmkg,hu2022empowering,jiang2023unikgqa,zhang2022greaselm} leverage LLMs for different KG-based tasks such as knowledge graph completion, graph-to-text generation, and question answering. The synergistic integration of LLMs and KGs ~\cite{feng2023knowledge,jiang2023structgpt,sun2023think} enables bidirectional reasoning grounded in both data and structured knowledge, as the two systems collaborate symbiotically to mutually enhance each other's capabilities.

While there have been studies on integrating knowledge graphs with LLMs, most of these focus on single-task scenarios. The potential of combining LLMs and knowledge graphs to enhance knowledge preservation in continual learning scenarios remains unexplored.

%% file: sec/3_method.tex
\section{Method}

In this section, we begin by introducing the setup of class-incremental learning and revisiting the training and testing procedures of GMM~\cite{cao2024GMM}. Then, we present our method from two perspectives: knowledge graph enhanced learning and knowledge graph augmented inference.

\subsection{Preliminaries}
\paragraph{Class Incremental Learning (CIL).} CIL is a paradigm that focuses on the progressive acquisition of knowledge across disjoint sets of classes. Let $\mathcal{X}$ denote the input space and $\mathcal{Y}$ represent the label space. At each incremental time step $t$, the learning algorithm receives a new dataset $D_t = \{(x_i, y_i) \mid x_i \in \mathcal{X}_t, y_i \in \mathcal{Y}_t\}$, where $\mathcal{Y}_t $ is a set of novel classes such that $\mathcal{Y}_t \cap (\cup_{k=1}^{t-1} \mathcal{Y}_k) = \emptyset$. The objective is to construct a classifier $f_t: \mathcal{X} \rightarrow \bigcup_{k=1}^t \mathcal{Y}_k$ that accurately predicts labels overall observed classes up to time $t$.

\begin{figure}
\centerline{\includegraphics[width=0.48 \textwidth]{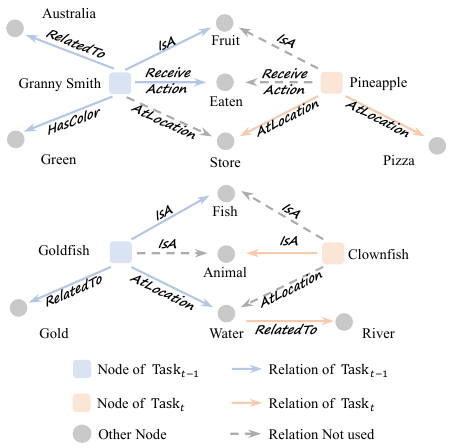}}
\caption{The knowledge graph construction process. Rectangle nodes represent class nodes, while round ones represent non-class nodes. Blue rectangles represent classes encountered in task $t-1$, while the orange ones represent classes from task $t$. Blue Arrows represent relations used for learning task $t-1$, while the orange ones represent relations used for task $t$. }

\label{fig:graph_construct}
\vspace{-5mm}
\end{figure}

\paragraph{Generative Multi-modal Models for CIL.}
In order to effectively harness the textual information embedded within the graph, we need to leverage language models capable of extracting text embeddings that can be aligned with image features. Currently, two continual learning frameworks can incorporate textual features: those based on CLIP~\cite{jha2024clap4clip, liu2023clipCIL,huang2024rapf} and those utilizing large language models (LLMs) ~\cite{cao2024GMM}. Given our need for more nuanced text understanding and encoding-decoding abilities, we adhere to the continual learning framework outlined in GMM~\cite{cao2024GMM}.
\begin{figure*}
\centerline{\includegraphics[width=0.999\textwidth]{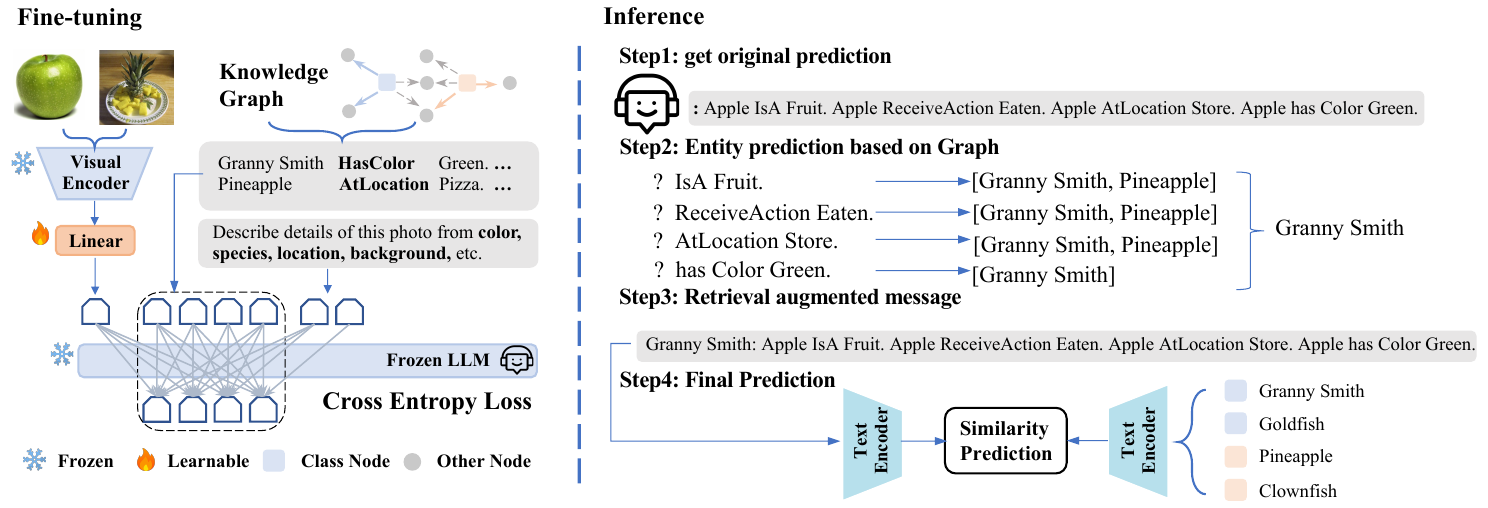}}
\caption{Left: In knowledge graph-enhanced learning, image embedding, knowledge-enhanced ground-truth embedding, and question embedding are input to Frozen LLM to generate predicted embedding; cross-entropy loss updates the linear layer. Right: In knowledge graph-augmented inference, relations from predicted text are extracted to create graph-augmented text, producing the final prediction.}
\label{fig:pipeline}
\end{figure*}

The GMM method’s training pipeline involves using a frozen image encoder  $f_{\text{enc}}$  and a text encoder  $f_{\text{text}}$   to process the image-text pairs 
$ \{ (X_{t,i}, S_{t,i}) \}_{i=1}^{N_t}$ for each task $t$, where $N_t$ means the total number of image-text pairs in task $t$ and $S_t$ is the text labels of classes in task $t$.
For each image $x_i$,  $f_{\text{enc}}$  generates an image embedding  $\mathbf{e}_i = f_{\text{enc}}(x_i; \theta_{\text{enc}})$, and for each class, GMM utilize the BERT tokenizer to get the corresponding embedding $\mathbf{s}_i$. Then, a question embedding $\mathbf{q}$ is also computed by the BERT model from the question text $q$. The image embedding $\mathbf{e}_i$ concatenated with the  ground-truth embedding $\mathbf{s}_i$ and question embedding $\mathbf{q}$, is input to a frozen LLM to get the predicted tokens sequentially based on prior tokens:
\begin{equation}
    P(\hat{\mathbf{s}}_1,..., \hat{\mathbf{s}}_{m}|{x}_i, \mathbf{{q}}, \mathbf{{s}}) = \prod \limits_{j=1}^{m-1}P(\mathbf{s_j}|\mathbf{{e}}_i, \mathbf{{q}}, \mathbf{s_1}, \ldots,\mathbf{s_{j-1}}).
\end{equation}

Then GMM computes Cross Entropy Loss to make the predicted token close to the ground-truth token:
\begin{equation}
\label{eq:cross_entropy}
    L_{\text{CE}} = -\frac{1}{m} \sum_{j=1}^m \mathbf{s}_j \cdot \log \mathbf{\hat{s}}_j,
\end{equation}
where  $\textbf{s}_j$  is the ground truth and  $\hat{\textbf{s}}_j$  is the prediction.

At inference time, GMM uses the fine-tuned model to get the text prediction based on the test image and instruction question. Then a text encoder  $f_{\text{text}}$  is used to measure the cosine similarity between predicted labels and ground truth to determine the final classification as: 
\begin{equation}
    \text{pred} = \arg\max \langle f_{\text{text}}(\mathbf{s}), f_{\text{text}}(\mathbf{\hat{s}}) \rangle.
\end{equation}

As discussed in the introduction, GMM may misclassify a predicted text to its word-related class label instead of its meaning-related ground truth (\eg, apple to pineapple instead of apple to Granny Smith). So we propose to use relations in a common sense graph to instruct the model output relation-aware text, thus helping locate the predicted text to its ground truth. We begin by introducing the Knowledge graph construction process, followed by how to utilize this graph for training and inference.

\subsection{Knowledge Graph Enhanced Learning}
\label{graph_sec}

\begin{algorithm}[ht]
\caption{Graph Construction For Task $t$}
\label{alg:graph}
       \textbf{Input:} $S_t$ \Comment{Class names of task $t$}\\
        \textbf{require:} $G = \{E, R, F\}$    \Comment{Entire Knowledge Graph} \\
        \textbf{require:} $E_{t-1}$  \Comment{Entity node for task $t-1$} \\
        \textbf{require:}  $G_{t-1}$  \Comment{Subgraph for task $t-1$} 
\begin{algorithmic}[1]

\State  $ E_t = E_{t-1} \cup S_t \cup \left\{ e \,|\, e \in \mathcal{T}(S_t) \right\} $ \Comment{Init $E_t$}
\State  $G'_t=\{E_{t}, R_{t-1},F_{t-1}\} $ \Comment{Init subgraph for task $t$}
\For {entity $e$ in $S_t$} \Comment{Iterate each new classes}
    \For {$(h, r, o)$ in $F$}  \Comment{Find unused facts}
    \If{$h == e$ and $(h,r,o) \notin F_{t-1}$ }
    \State $G'_t = \{E_t, R_t \cup r, F_t \cup (h,r,o)\}$
    \EndIf

    \EndFor
\EndFor

\State \textbf{return} $G'_t$ \Comment{Return the subgraph for task $t$}
\end{algorithmic}

\end{algorithm}

\minisection{Knowledge Graph Construction.} 
In order to extract relations for each encountered new category, we follow ZSL-KG~\cite{nayaktmlr22} to query three tables: nodes, relations, and edges from the ConceptNet database based on the ILSVRC-21K~\cite{imgnet} classes within two-hop relations.

Then, we have a common sense knowledge graph at hand presented as $G = \{E, R, F\}$, where $E$ represents entities, $R$ represents   relations and $F=\{(h, r, o)\} \subseteq E \times R \times E$ represents facts contained in this graph, in which $h$ means head entity and $o$ means tail object. In the continual learning process, we gradually build a dataset-specific graph in the order in which classes appear at each given task.

To better illustrate our method, we divide our graph construction process into two steps:

\minisection{Step 1: Build a temporary knowledge graph.}
Suppose we are currently in task $t$ of continual learning. We initialize 
\begin{equation}
    E_t = E_{t-1} \cup S_t \cup \left\{ e \,|\, e \in \mathcal{T}(S_t) \right\},
\end{equation}
where $S_t$ denotes new class nodes, and $\mathcal{T}(S_t)$ extracts non-class nodes from relation triplets involving $S_t$. Then we build a temporary graph for task $t$:
\begin{equation}
    G_t = (E_t, R_t, F_t),
\end{equation}
containing all entities, relations, and fact triplets observed up to task $t$.

\minisection{Step 2: Trim to Key Relationships.}
To enable efficient training and improved discrimination between similar classes, we proposed to trim the temporary knowledge graph $G_t$ to a smaller one $G'_t$:
\begin{equation}
    G'_t = (E'_t, R'_t, F'_t),
\end{equation}
that only contains a few unique relations that best distinguish it from similar old classes. Here, $E'_t$ is the trimmed entity set:
\begin{equation}
    E_{t}^{'} = \left\{ e \mid e \in E_t,\ e \notin \bigcup_{k=1}^{t-1} E_{k}^{'} \right\}.
\end{equation}
$R'_t$ and $F'_t$ are relations and facts that are associated with the current entity set $G'_t$. For example, as illustrated in Fig.~\ref{fig:graph_construct}, we choose ``AtLocation Store'' and ``AtLocation Pizza'' instead of previously used ``IsA Fruit'' and ``ReceiveAction Eaten'' by class ``Granny Smith'' from task $t-1$. The whole construction process of $G'_t$ is detailed in Algo.~\ref{alg:graph} Lines 4-6.

There exist circumstances where classes from later tasks (maybe the last task) have no direct relation to use; that is, all directly related tail nodes have been taken by former encountered classes. We tackle this problem by using relations of the second hop in the common sense knowledge graph $G$ (\eg, we use ``Clownfish ReatedTo Water RelatedTo River'' instead of the previously occupied node ``Water''). 

After obtaining the distinct graph that has different relations for each class, we then concatenate $m$ triplets of each class together and input them into the model as the ground truth text $s_i$ in the image-text pairs as shown in the left side of Fig.~\ref{fig:pipeline}, fine-tuning the linear layer after the image encoder with Cross Entropy loss in Eq.~\ref{eq:cross_entropy}.

\subsection{Knowledge Graph Augmented Inference}
The inference process is illustrated on the right side of Fig.~\ref{fig:pipeline}. 
We first get the relation-aware output by the fine-tuned LLM with instructing questions ``Describe details of this photo from color, species, location, background, etc.''.
Then we decompose the raw output  $s$  into  $m$  triplets: $\{ (h_p, r_p, o_p) \}_{p=1}^m$.

After obtaining the predicted relations, we remove the head entities $h_p$ and use the relation pairs  $\{(r_p, o_p)\}_{p=1}^m$  to search within the subgraph  $G_t$  constructed for the current task. The head entity that appears most frequently in the search results is considered the predicted head entity:
\begin{equation}
    \hat{h} = \arg\max_{h} \sum_{p=1}^m \mathbb{I}\left[ (h, r_p, o_p) \in G_t \right],
\end{equation}
where $\mathbb{I}[\cdot]$ is the indicator function that equals $1$ if the triplet is correct ( $(h, r_p, o_p) \in E_t$) and $0$ otherwise.

Although this graph-based prediction $\hat{h}$ can be treated as the final prediction, it neglects the rich information contained in the original raw text output $s$. To leverage both the knowledge reservation ability provided by the graph and the text understanding and reasoning ability provided by LLM, we prepend this graph-based prediction $\hat{h}$ to  $s$  to obtain the graph-augmented output:
\begin{equation}    
    s_a = \hat{h} \, \oplus \, s,
\end{equation}
where $\oplus$ denotes the concatenation operation, meaning that $s_a$ is formed by placing $\hat{h}$ directly before $s$.
This augmented sentence  $s_a$  is then input into the text encoder to perform similarity prediction between the encoded  $s_a$  and all the class features encountered so far to obtain the final prediction result:
\begin{equation}
    pred = \arg\max_{c \in S_t} \, \text{sim}\left( f_{text}(s_a), f_{text}(c) \right).
\end{equation}

\begin{table*}
 \caption{The performance comparison results between our method and other methods on Tiny-ImageNet and ImageNet-R.}
    \centering
    \resizebox{0.95\textwidth}{!}{%
    \begin{tabular}{llcccccccc}
    \toprule
    \multirow{3}{*}{Type}                                                       & \multirow{3}{*}{Method}       & \multirow{3}{*}{\begin{tabular}[c]{@{}c@{}}Exemplar\end{tabular}} & \multicolumn{6}{c}{Tiny-ImageNet}                                                                                                       & ImageNet-R           \\ \cline{4-10}  
                                                                                &                             &                                                                          & \multicolumn{2}{c}{B100-5 tasks}                 & \multicolumn{2}{c}{B100-10 tasks}                & \multicolumn{2}{c}{B100-20 tasks}                & B0-10 tasks             \\
                                                                                &                             &                                                                          & \textbf{Avg}         & \textbf{Last}        & \textbf{Avg}         & \textbf{Last}        & \textbf{Avg}         & \textbf{Last}        & Last                 \\ \hline
    \multirow{7}{*}{Conventional}                                                & EWC~\cite{kirkpatrick2017overcoming}~                        & \XSolidBrush                                                                         & 19.01                & 6.00                 & 15.82                & 3.79                 & 12.35                & 4.73                 & 35.00                 \\
                                                                                & LwF~\cite{li2017learning}          &  \XSolidBrush                                                                        & 22.31                & 7.34                 & 17.34                & 4.73                 & 12.48                & 4.26                 & 38.50                 \\
                                                                                & iCaRL~\cite{rebuffi2017icarl}  & \Checkmark                                                                         & 45.95                & 34.60                & 43.22                & 33.22                & 37.85                & 27.54                & -                    \\
                                                                                & EEIL~\cite{castro2018end}          &  \Checkmark                                                                        & 47.17                & 35.12                & 45.03                & 34.64                & 40.41                & 29.72                & -                    \\
                                                                                & UCIR~\cite{UCIR_2019_CVPR}        &     \Checkmark                                                                     & 50.30                & 39.42                & 48.58                & 37.29                & 42.84                & 30.85                & -                    \\
                                                                                & PASS~\cite{zhu2021prototype}       &   \XSolidBrush                                                                       & 49.54                & 41.64                & 47.19                & 39.27                & 42.01                & 32.93                & -                    \\
                                                                                & DyTox~\cite{douillard2021dytox} & \Checkmark                                                                         & 55.58                & 47.23                & 52.26                & 42.79                & 46.18                & 36.21                & -                    \\ \hline
    \multirow{7}{*}{\begin{tabular}[c]{@{}l@{}}Discriminative \\PT models\end{tabular}}   & Continual-CLIP\cite{thengane2022continualclip}              & \XSolidBrush                                                                         & 70.49       & 66.43      & 70.55     & 66.43       & 70.51       & 66.43       & 72.00                     \\
                                                                                & L2P~\cite{l2p}                         &  \XSolidBrush                                                                         &          83.53            &        78.32              &        76.37              &          65.78            &       68.04               &         52.40             &       72.92                \\
                                                                                & DualPrompt~\cite{wang2022dualprompt}                       &  \XSolidBrush                                                                         &      85.15                &    81.01                  &          81.38            &    73.73                  &    73.45                  &      60.16                & 68.82                     \\
                                                                                 & CODA-Prompt~\cite{smith2023coda}                         & \XSolidBrush                                                                         &             {85.91}         &         {81.36}             &  82.80                    &   75.28                   &          77.43            &  66.32                    &   73.88                   \\
                                                                                 & MoE-CLIP~\cite{yu2024boosting}                         & \XSolidBrush                                                                         &  81.12                     &  76.81                     & 80.23                     & 76.35                      & 79.96                      & 75.77                      &  80.87                     \\
                                                                                 & RAPF~\cite{huang2024class}                         & \XSolidBrush                                                                         &  78.64                     & 74.67                      & 77.42                      & 73.57                      & 76.29                      & 72.65                      &  80.28                     \\
                                                                                & Linear Probe                &      \XSolidBrush                                                                     &   74.38                   &          65.40            &         69.73             &     58.31                 &          60.14            &   49.72                   &     45.17                 \\
                                                                                \hline
    \multirow{3}{*}{\begin{tabular}[c]{@{}l@{}}Generative\\ PT models\end{tabular}} & Zero-shot                   &  \XSolidBrush                                                                         &       58.16               &       53.72               &    58.10                  &           53.72           &           58.13           &         53.72             &       67.38               \\
                                                                                & GMM~\cite{cao2024GMM}                         &  \XSolidBrush                                                                         & 83.42                     & 76.98                     &  82.49                    & 76.51                     & 81.70                     & 76.03                     &        80.72              \\
                                                                                & KG-GMM (Ours)                        &  \XSolidBrush                                                                       & \textbf{86.17}                   & \textbf{81.86}                 &  \textbf{84.37}                &  \textbf{78.16}                    &  \textbf{83.18}                 & \textbf{78.32}                  &  \textbf{84.29}  \\
                                                                                  
                                                                                
                                                                                \bottomrule
    \end{tabular}%
    }
   
    \label{tab:tiny_CIL}
    
    \end{table*}

%% file: sec/4_exp.tex
\section{Experiments}
\begin{table*}
\caption{Comparison results of our method with other conventional baselines and methods on the mini-ImageNet dataset for few-shot class incremental learning. The table includes one base task and eight incremental tasks. \textbf{PD} is the performance drop between the first and last session. $^{*}$ indicates our re-implementation based on PILOT~\cite{sun2023pilot}.}
\centering

\resizebox{0.9\textwidth}{!}{\begin{tblr}{
  hline{1} = {-}{0.08em},
  hline{2, 13} = {-}{},
  hline{16} = {-}{0.08em},
}
            & 0 & 1     & 2     & 3     & 4     & 5     & 6     & 7     & 8     & PD↓ & HAcc↑   \\
iCaRL~\cite{rebuffi2017icarl}       & 61.31   & 46.32 & 42.94 & 37.63 & 30.49 & 24.00    & 20.89 & 18.80  & 17.21 & 44.10&32.45  \\
EEIL~\cite{castro2018end}        & 61.31   & 46.58 & 44.00    & 37.29 & 33.14 & 27.12 & 24.10  & 21.57 & 19.58 & 41.73 &28.43\\
LUCIR~\cite{UCIR_2019_CVPR}       & 61.31   & 47.80  & 39.31 & 31.91 & 25.68 & 21.35 & 18.67 & 17.24 & 14.17 & 47.14 &35.65\\
TOPIC~\cite{tao2020few}       & 61.31   & 50.09 & 45.17 & 41.16 & 37.48 & 35.52 & 32.19 & 29.46 & 24.42 & 36.89 &32.98\\
CEC~\cite{zhang2021few}         & 72.00      & 66.83 & 62.97 & 59.43 & 56.70  & 53.73 & 51.19 & 49.24 & 47.63 & 24.37 &15.96\\
F2M~\cite{shi2021overcoming}         & 72.05   & 67.47 & 63.16 & 59.70  & 56.71 & 53.77 & 51.11 & 49.21 & 47.84 & 24.21 &19.21\\
MetaFSCIL~\cite{chi2022metafscil}   & 72.04   & 67.94 & 63.77 & 60.29 & 57.58 & 55.16 & 52.90  & 50.79 & 49.19 & 22.85 &14.35\\
Entropy-reg~\cite{liu2022few} & 71.84   & 67.12 & 63.21 & 59.77 & 57.01 & 53.95 & 51.55 & 49.52 & 48.21 & 23.63 &19.29\\
L2P$^{*}$~\cite{l2p} & 94.12   &87.20 & 80.99 & 75.67 & 70.94 & 66.76 & 63.11 & 59.81 & 56.83 &  37.29&0.00\\
DualPrompt$^{*}$~\cite{wang2022dualprompt} & 93.97   & 86.85 & 80.67 & 75.31 & 70.61 & 66.44 & 62.77 & 59.58 & 56.80 & 37.17 &0.10 \\
CODA-Prompt$^{*}$~\cite{smith2023coda} & \textbf{95.37}   & 88.86 & 82.69 & 77.87 & 74.47 & 70.16 & 66.46 & 63.73 & 61.14 & 34.23 &0.00\\
Zero-shot & 58.08   & 58.95 & 57.76 & 57.89 & 58.19 & 57.42 & 56.26 & 54.82 & 54.95 & \textbf{3.13} &52.47\\
GMM        &  89.35  & 88.40 & 86.11 &85.07  &83.61 & 81.35 & 78.97 & 77.34 & 75.18 & 14.17&71.45 \\
KG-GMM (Ours)        &  90.99  & \textbf{89.50} & \textbf{88.11} &\textbf{87.29}  &\textbf{85.63} & \textbf{83.70} & \textbf{81.12} & \textbf{79.63} & \textbf{78.07} & 12.93 &\textbf{74.81}
\end{tblr}}

\label{tab:mini_few}
\end{table*}

\subsection{Experiments Setup}
\minisection{Datasets.}
%
We test our model on two commonly used continual learning benchmarks: Tiny-ImageNet and ImageNet-R, and two few-shot continual learning benchmarks: CIFAR100 and Mini-ImageNet.

For conventional continual learning, we follow the two standard configurations used in GMM~\cite{cao2024GMM}: B0, in which all classes are equally divided among different tasks, and B100 (i.e. Tiny-ImageNet) in which the first task contains 100 classes (half of the dataset) and the rest are equally divided into subsequent tasks. For few-shot continual learning, we follow the data splits proposed by \cite{tao2020few}. For both datasets, we divide the data into two parts: a base session and incremental sessions. The base session consists of 60 classes with full access to all associated data. Each incremental session follows a 5-way 5-shot setting, introducing 5 new classes with only 5 samples per class.

We build our expanding knowledge graph based on ConceptNet, which is a large-scale, multilingual knowledge graph that represents common-sense relationships between words and phrases in natural language. It consists of over 8 million nodes and approximately 21 million edges, connecting concepts through 50 relationships including “IsA”, “PartOf”, “UsedFor” and “HasProperty”, etc.

\minisection{Implementation details.}
For the Knowledge Graph part, we follow ZSL-KG~\cite{nayaktmlr22} to use a 2-hop ImageNet-based knowledge graph extracted from ConceptNet~\cite{speer2017conceptnet} as the initial $G$ described in section~\ref{graph_sec}. $G$ contains $574270$ nodes $E$, $50$ relations $R$ and $1380131$ edges $F$.
For the Generative Multi-modal model part, we follow GMM to use the MiniGPT-4~\cite{zhu2023minigpt} framework as the image and text encoder. In the B0 setting of all datasets, we employ a 200-iteration warmup with a learning rate of 3e-6 and a learning rate from 3e-5 to 3e-6 with a cosine decay scheduler in the following fine-tuning phase. In the B100 setting, we first employ a learning rate of 3e-6, and then on the subsequent tasks, we adopt a lower learning rate of 3e-7, both employing a cosine decay scheduler.

\minisection{Baselines and evaluating metrics.}
We follow GMM~\cite{cao2024GMM} to compare with three different categories of methods, including conventional train-from-scratch based~\cite{rebuffi2017icarl, UCIR_2019_CVPR, zhao2020maintaining,
yan2021dynamically, douillard2021dytox,shi2021overcoming, li2017learning, castro2018end, zhu2021prototype}, discriminative pre-trained based~\cite{l2p,wang2022dualprompt,smith2023coda} and generative pre-trained based~\cite{cao2024GMM}. The evaluation metrics for the experiments are defined as follows: “Avg” represents the model’s average accuracy across all tasks, while “Last” denotes the model’s accuracy on all test sets after fine-tuning the final task. For few-shot continual learning, we also add a new metric Harmonic Accuracy~\cite{peng2022few} (HAcc) to check the balanced performance between the base and new classes. $A_H=\frac{2\times A_0\times A_n}{A_0 + A_n}$, where $A_0$ is the acc of base classes and $A_n$ is the average of all classes. We report an average accuracy of three runs based on three different class orders. Please refer to the Supplementary Material for detailed results of different orders.

\subsection{Experiments on Conventional CIL}

In Tab.~\ref{tab:tiny_CIL}, we present experiments on 100 base classes with 5, 10, and 20 incremental tasks settings on Tiny-ImageNet, and B0-10 tasks setting on ImageNet-R. Our KG-GMM demonstrates superior performance across all settings of these two datasets. On Tiny-ImageNet, our method consistently outperforms others in all settings, particularly we outperform GMM by $2.93\%$ on average in three settings in terms of Last accuracy.
On ImageNet-R, our model achieved an impressive last task performance of $84.29\%$, surpassing the previous SOTA GMM by $3.57\%$, the previous best prompt-based method CODA-Prompt by $10.41\%$.

\begin{table}[!]
\caption{Accuracy comparison under the few-shot class incremental learning on CIFAR-100. For simplicity, we present the accuracy after one base task, the fourth task, and the final eighth task.}
    \centering
    \resizebox{0.46\textwidth}{!}{\begin{tblr}{
      hline{1} = {-}{0.08em},
  hline{2, 13} = {-}{},
  hline{16} = {-}{0.08em},
    }
    Method      & 0 & 4     & 8     & PD↓ & HAcc↑  \\
    iCaRL~\cite{rebuffi2017icarl}       & 64.10     & 27.93 & 13.73 & 50.37 &41.45 \\
    EEIL~\cite{castro2018end}        & 64.10     & 28.96 & 15.85 & 48.25 &32.43\\
    LUCIR~\cite{UCIR_2019_CVPR}       & 64.10     & 31.61 & 13.54 & 50.56 & 39.67 \\
    TOPIC~\cite{tao2020few}       & 64.10     & 40.11 & 29.37 & 34.73& 25.23 \\
    CEC~\cite{zhang2021few}         & 73.07    & 58.09 & 49.14 & 23.93&22.46 \\
    F2M~\cite{shi2021overcoming}         & 71.45    & 57.76 & 49.35 & 22.06 &19.37 \\
    MetaFSCIL~\cite{chi2022metafscil}   & 74.50     & 59.48 & 49.97 & 24.53&3.60 \\
    Entropy-reg~\cite{liu2022few} & 74.40     & 59.71 & 50.14 & 24.26&11.53 \\
    L2P$^{*}$~\cite{l2p}         & 91.22    & 68.66 & 54.89 & 36.33 &0.00 \\
    DualPrompt$^{*}$~\cite{wang2022dualprompt}  & 91.08    & 68.45 & 54.67 & 36.41&0.10 \\
    CODA-Prompt$^{*}$~\cite{smith2023coda} & \textbf{93.55}    & 71.91 & 59.32 & 34.23&0.00 \\
    Zero-shot & 74.13    & 72.59 & 67.93 & \textbf{6.20}&64.75 \\
    GMM~\cite{cao2024GMM}         & 91.53    & 85.65 & 81.47 & 10.06 &75.43\\
    KG-GMM (Ours)       & 91.83    & \textbf{86.23} & \textbf{82.37} & 9.46&\textbf{79.68} 
    \end{tblr}}
    \label{tab:cifar_few}
    \vspace{-5mm}
    \end{table} 

\subsection{Experiments on Few-shot CIL}

In Tab.~\ref{tab:mini_few}, we present the comparison results of our method against other baselines on Mini-ImageNet for few-shot class incremental learning. The table shows that our model lags behind several ImageNet-21K pre-trained methods in the base session, it achieves state-of-the-art results in all subsequent eight sessions. Specifically, in the final session, our method reaches an accuracy of $78.07\%$, outperforming DualPrompt~\cite{wang2022dualprompt} by $21.26\%$ points, CODA-Prompt~\cite{smith2023coda} by $16.93\%$ points, and the previous best GMM~\cite{cao2024GMM} by $2.89\%$ points. Notably, our model leads GMM by $1.64\%$ points in the first incremental session and extends this lead to $2.89$ points in the final session, indicating that our approach more effectively mitigates forgetting through the integration of LLM and knowledge graphs.

In Tab.~\ref{tab:cifar_few}, we present the comparison results of our method against other baselines on CIFAR100 for few-shot class incremental learning. The results indicate that our method also achieves superior performance on low-resolution datasets, outperforming the previous state-of-the-art method GMM by $0.90\%$ points, traditional train-from-scratch method Entropy-reg~\cite{liu2022few} by $32.23\%$ points, and the state-of-the-art prompt-based method CODA-Prompt~\cite{smith2023coda} by $23.05\%$ points in the final session. Note that although our method shows only a slight improvement over GMM in the last session, it achieves a $4.25\%$ gain in Harmonic Accuracy. This indicates that our approach better balances learning new classes while retaining previously learned ones.

\subsection{Further Analysis}
\minisection{Ablation study on different components.}
In Tab.~\ref{tab:ab}, we present ablation studies to validate the effectiveness of our method. We first use the GPT-3.5 generated descriptions (using the same prompt question as our method) for each class as the text in the image-text pair of the original GMM. The results show that without the explicit guidance of the relation keywords, the rich information brought by the text has almost no improvement over the simple text labels used in the original GMM.  Besides, we show that using only the head entity $\hat{h}$ provided by the search results in KG (third row in Tab.~\ref{tab:ab} shows considerable improvement ($82.77\%$), but combined with the text output, our proposed KG-GMM achieves the best results $84.29\%$.

\begin{table}
\caption{The ablation study results on ImageNet-R B0-10 tasks. GMM + Descriptions Labels: only use class descriptions generated by GPT-3.5 as data augmentation. KG-GMM wo Graph-Augmented Inference: using graph-based prediction $\hat{h}$ as the final prediction. KG-GMM: our method.}
\centering
\resizebox{0.42\textwidth}{!}{
\begin{tblr}{
  hline{1-2,6} = {-}{},
}
Method                                          & Accuracy  \\
GMM                                           & 80.72 \\
GMM+Descriptions Labels                       & 80.95 \\
KG-GMM w/o Graph-Augmented Inference                             & 82.77 \\
KG-GMM  & 84.29 
\end{tblr}}

\label{tab:ab}
\vspace{-5mm}
\end{table}

\begin{table}[!]
\caption{Time complexity analysis on the last task of B0-10 tasks setting of the ImageNet-R dataset.}
\centering
\resizebox{0.42\textwidth}{!}{
\begin{tabular}{lllll}
\toprule
Relation Stored     & r = 0   & r = 2   & r = 3 (Ours) & r=4  \\ \midrule
Avg text length (c)  & 30    & 51    & 75  & 102  \\
Generation cost (s)  & 6.54  & 7.42  & 9.67  &12.63 \\
Storage (MB)         & 0     & 0.05  & 0.07  & 0.08 \\
Graph Inference (ms) & 0     & 0.46  & 0.53  & 0.62\\
Accuracy (\%)        & 81.03 & 81.92 & 84.29 & 84.31\\ \bottomrule
\end{tabular}
}
\label{choosingr}
\vspace{-2mm}
\end{table}

    \begin{figure*}[!]
    \centering
    \includegraphics[width=0.99\textwidth]
    {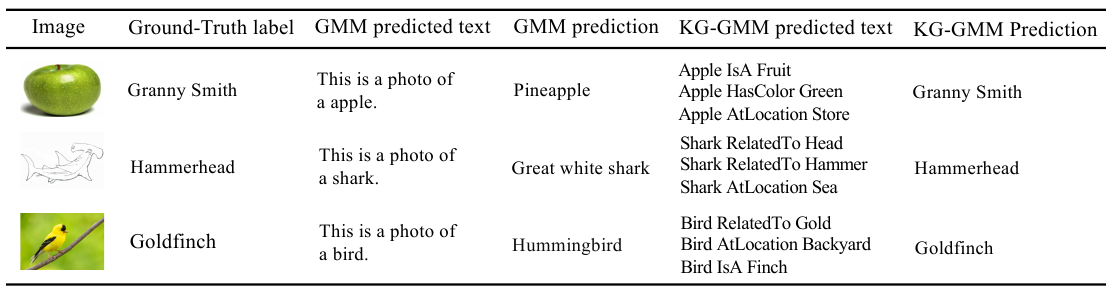}
    \vspace{-2mm}
    \caption{Text examples of our methods against the original GMM.}
    \label{fig:vis}
    \vspace{-4mm}
    \end{figure*}

\begin{figure}[t]
    \centering
    \includegraphics[width=0.44\textwidth]
    {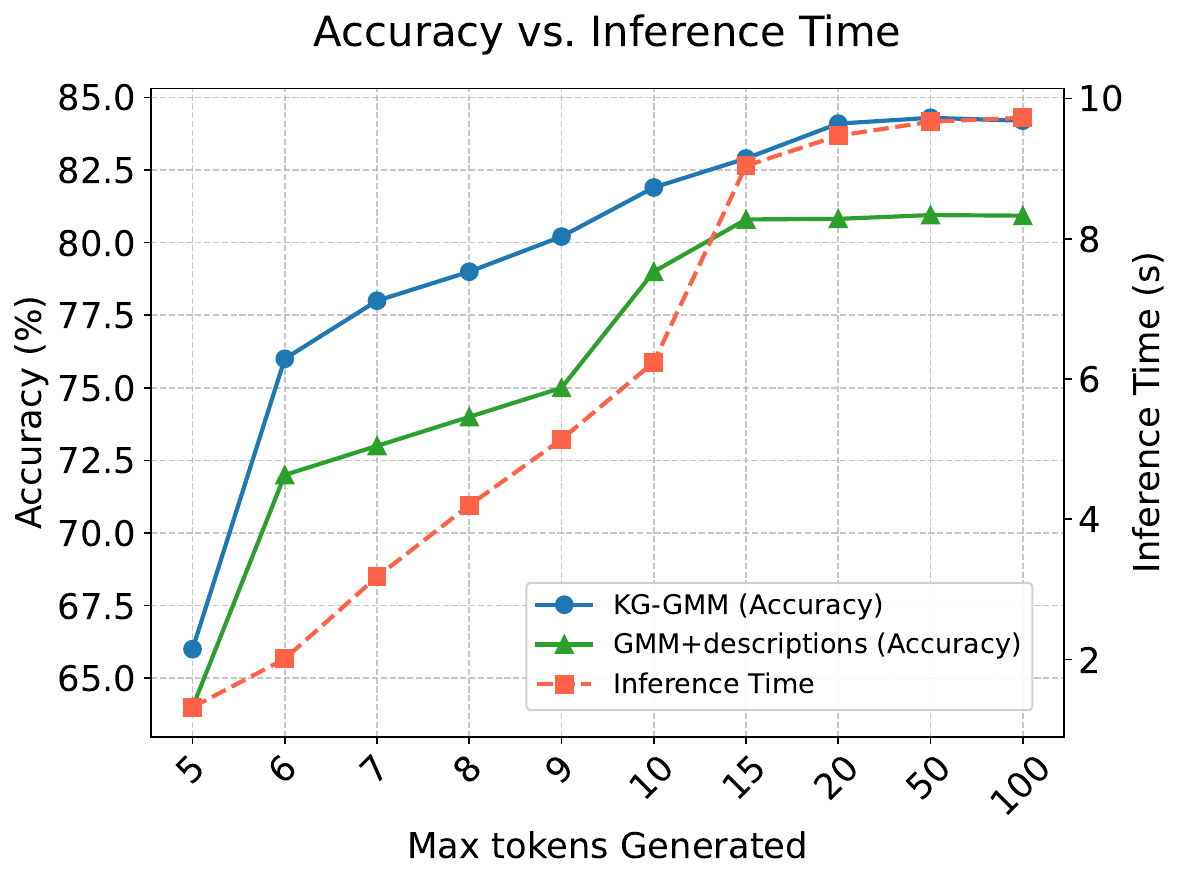}
   
    \caption{Model Accuracy vs. Inference Time comparison regarding max tokens configured during inference.
    }
    \label{time_tokens}
    \vspace{-5mm}
    \end{figure}

\minisection{Ablation on Max tokens.} In Fig.~\ref{time_tokens} we present the comparative analysis of maximum token limits on model performance and inference time. The experiments were conducted on ImageNet-R, comparing our KG-GMM method with the baseline approach GMM+Descriptions that utilizes class descriptions generated by GPT-3.5 for data augmentation. Inference time was measured as the averaged processing duration per batch (size=64).

Notably, our method achieves near-optimal performance at 20 tokens, while GMM+Descriptions exhibits slower performance growth with increasing token limits. This indicates that the knowledge graph-augmented GMM produces more information-dense outputs at lower token constraints, demonstrating two key advantages: 1) Enhanced information efficiency through KG-enhanced training enables effective knowledge condensation, and 2) Superior performance is attained with reduced computational overhead.

\minisection{Time complexity analysis.}
In Tab.~\ref{choosingr}, we present the additional storage and inference costs associated with selecting different numbers of relations $r$ per class with a total batch size of 64.
For $r = 0$, we used GMM's default prompt, ``This is a photo of [CLS]'', while for $r=2$ and $r=3$, the average generated text lengths correspond to 51 and 75 characters, respectively. ``Generation cost'' represents the time required to generate text, measured in seconds, ``Storage'' denotes the additional memory usage from the knowledge graph, measured in MB, and ``Graph inference'' indicates the time needed to obtain the graph-based prediction $\hat{h}$, measured in milliseconds.

As shown in Tab.~\ref{choosingr}, the storage and inference overhead of the knowledge graph is minimal ($0.07$MB and $0.53$ms for $r=3$). The primary cost arises during the inference phase (generating longer texts increases inference time with higher values of $r$). Balancing inference time with performance gains, we select $r=3$ as our final hyperparameter.

\minisection{Illustration of corrected predictions.}
We present some visual examples of our proposed KG-GMM in terms of the predicted text and final prediction in the B0-10 tasks setting of the ImageNet-R dataset. We can see that the GMM can still recognize a shark but lose the ability to classify it to the right subclass (hammerhead). Instead, our KG-GMM can extract rich information from the relation contained in the predicted text and locate it to the right class. 
We refer the readers to the supplementary materials for more visualization results including failure cases.

%% file: main.bbl
\begin{thebibliography}{64}
\providecommand{\natexlab}[1]{#1}
\providecommand{\url}[1]{\texttt{#1}}
\expandafter\ifx\csname urlstyle\endcsname\relax
  \providecommand{\doi}[1]{doi: #1}\else
  \providecommand{\doi}{doi: \begingroup \urlstyle{rm}\Url}\fi

\bibitem[Aljundi et~al.(2018)Aljundi, Babiloni, Elhoseiny, Rohrbach, and Tuytelaars]{aljundi2018memory}
Rahaf Aljundi, Francesca Babiloni, Mohamed Elhoseiny, Marcus Rohrbach, and Tinne Tuytelaars.
\newblock Memory aware synapses: Learning what (not) to forget.
\newblock In \emph{Proceedings of the European conference on computer vision (ECCV)}, pages 139--154, 2018.

\bibitem[Bodenreider(2004)]{bodenreider2004unified}
Olivier Bodenreider.
\newblock The unified medical language system (umls): integrating biomedical terminology.
\newblock \emph{Nucleic acids research}, 32\penalty0 (suppl\_1):\penalty0 D267--D270, 2004.

\bibitem[Cao and Liu(2022)]{cao2022relmkg}
Xing Cao and Yun Liu.
\newblock Relmkg: reasoning with pre-trained language models and knowledge graphs for complex question answering.
\newblock \emph{Applied Intelligence}, pages 1--15, 2022.

\bibitem[Cao et~al.(2024)Cao, Lu, Huang, Liu, and Cheng]{cao2024GMM}
Xusheng Cao, Haori Lu, Linlan Huang, Xialei Liu, and Ming-Ming Cheng.
\newblock Generative multi-modal models are good class incremental learners.
\newblock \emph{IEEE Computer Vision and Pattern Recognition (CVPR)}, 2024.

\bibitem[Castro et~al.(2018)Castro, Mar{\'\i}n-Jim{\'e}nez, Guil, Schmid, and Alahari]{castro2018end}
Francisco~M Castro, Manuel~J Mar{\'\i}n-Jim{\'e}nez, Nicol{\'a}s Guil, Cordelia Schmid, and Karteek Alahari.
\newblock End-to-end incremental learning.
\newblock In \emph{Proceedings of the European conference on computer vision (ECCV)}, pages 233--248, 2018.

\bibitem[Chaudhry et~al.(2018)Chaudhry, Dokania, Ajanthan, and Torr]{Chaudhry_2018_ECCV}
Arslan Chaudhry, Puneet~K. Dokania, Thalaiyasingam Ajanthan, and Philip H.~S. Torr.
\newblock Riemannian walk for incremental learning: Understanding forgetting and intransigence.
\newblock In \emph{Proceedings of the European Conference on Computer Vision (ECCV)}, 2018.

\bibitem[Chi et~al.(2022)Chi, Gu, Liu, Wang, Yu, and Tang]{chi2022metafscil}
Zhixiang Chi, Li Gu, Huan Liu, Yang Wang, Yuanhao Yu, and Jin Tang.
\newblock Metafscil: A meta-learning approach for few-shot class incremental learning.
\newblock In \emph{Proceedings of the IEEE/CVF conference on computer vision and pattern recognition}, pages 14166--14175, 2022.

\bibitem[Deng et~al.(2009)Deng, Dong, Socher, Li, Li, and Fei-Fei]{imgnet}
Jia Deng, Wei Dong, Richard Socher, Li-Jia Li, Kai Li, and Li Fei-Fei.
\newblock Imagenet: A large-scale hierarchical image database.
\newblock In \emph{2009 IEEE Conference on Computer Vision and Pattern Recognition}, pages 248--255, 2009.

\bibitem[Douillard et~al.(2022)Douillard, Ram\'e, Couairon, and Cord]{douillard2021dytox}
Arthur Douillard, Alexandre Ram\'e, Guillaume Couairon, and Matthieu Cord.
\newblock Dytox: Transformers for continual learning with dynamic token expansion.
\newblock In \emph{Proceedings of the IEEE Conference on Computer Vision and Pattern Recognition (CVPR)}, 2022.

\bibitem[Feng et~al.(2023)Feng, Zhang, and Fei]{feng2023knowledge}
Chao Feng, Xinyu Zhang, and Zichu Fei.
\newblock Knowledge solver: Teaching llms to search for domain knowledge from knowledge graphs.
\newblock \emph{arXiv preprint arXiv:2309.03118}, 2023.

\bibitem[Gao et~al.(2024)Gao, Dong, He, Wang, and Gong]{2024beyond}
Xinyuan Gao, Songlin Dong, Yuhang He, Qiang Wang, and Yihong Gong.
\newblock Beyond prompt learning: Continual adapter for efficient rehearsal-free continual learning.
\newblock In \emph{ECCV}, 2024.

\bibitem[Gomez-Villa et~al.(2024)Gomez-Villa, Goswami, Wang, Andrew, Twardowski, and van~de Weijer]{2024LDC}
Alex Gomez-Villa, Dipam Goswami, Kai Wang, Bagdanov Andrew, Bartlomiej Twardowski, and Joost van~de Weijer.
\newblock Exemplar-free continual representation learning via learnable drift compensation.
\newblock In \emph{European Conference on Computer Vision}, 2024.

\bibitem[Hou et~al.(2019)Hou, Pan, Loy, Wang, and Lin]{UCIR_2019_CVPR}
Saihui Hou, Xinyu Pan, Chen~Change Loy, Zilei Wang, and Dahua Lin.
\newblock Learning a unified classifier incrementally via rebalancing.
\newblock In \emph{CVPR}, 2019.

\bibitem[Hu et~al.(2022)Hu, Xu, Yu, Wang, Yang, Zhu, Chang, and Sun]{hu2022empowering}
Ziniu Hu, Yichong Xu, Wenhao Yu, Shuohang Wang, Ziyi Yang, Chenguang Zhu, Kai-Wei Chang, and Yizhou Sun.
\newblock Empowering language models with knowledge graph reasoning for open-domain question answering.
\newblock In \emph{EMNLP}, pages 9562--9581, 2022.

\bibitem[Huang et~al.(2024{\natexlab{a}})Huang, Cao, Lu, and Liu]{huang2024class}
Linlan Huang, Xusheng Cao, Haori Lu, and Xialei Liu.
\newblock Class-incremental learning with clip: Adaptive representation adjustment and parameter fusion.
\newblock In \emph{European Conference on Computer Vision}, pages 214--231. Springer, 2024{\natexlab{a}}.

\bibitem[Huang et~al.(2024{\natexlab{b}})Huang, Cao, Lu, and Liu]{huang2024rapf}
Linlan Huang, Xusheng Cao, Haori Lu, and Xialei Liu.
\newblock Class-incremental learning with clip: Adaptive representation adjustment and parameter fusion.
\newblock In \emph{ECCV}, 2024{\natexlab{b}}.

\bibitem[Ilievski et~al.(2021)Ilievski, Szekely, and Zhang]{ilievski2021cskg}
Filip Ilievski, Pedro Szekely, and Bin Zhang.
\newblock Cskg: The commonsense knowledge graph.
\newblock In \emph{The Semantic Web: 18th International Conference, ESWC 2021, Virtual Event, June 6--10, 2021, Proceedings 18}, pages 680--696. Springer, 2021.

\bibitem[Jha et~al.(2024)Jha, Gong, and Yao]{jha2024clap4clip}
Saurav Jha, Dong Gong, and Lina Yao.
\newblock {CLAP4CLIP}: Continual learning with probabilistic finetuning for vision-language models.
\newblock In \emph{Thirty-eighth Conference on Neural Information Processing Systems}, 2024.

\bibitem[Jia et~al.(2021)Jia, Yang, Xia, Chen, Parekh, Pham, Le, Sung, Li, and Duerig]{jia2021scaling}
Chao Jia, Yinfei Yang, Ye Xia, Yi-Ting Chen, Zarana Parekh, Hieu Pham, Quoc Le, Yun-Hsuan Sung, Zhen Li, and Tom Duerig.
\newblock Scaling up visual and vision-language representation learning with noisy text supervision.
\newblock In \emph{International conference on machine learning}, pages 4904--4916. PMLR, 2021.

\bibitem[Jiang et~al.(2023{\natexlab{a}})Jiang, Zhou, Dong, Ye, Zhao, and Wen]{jiang2023structgpt}
Jinhao Jiang, Kun Zhou, Zican Dong, Keming Ye, Wayne~Xin Zhao, and Ji-Rong Wen.
\newblock Structgpt: A general framework for large language model to reason over structured data.
\newblock \emph{arXiv preprint arXiv:2305.09645}, 2023{\natexlab{a}}.

\bibitem[Jiang et~al.(2023{\natexlab{b}})Jiang, Zhou, Zhao, and Wen]{jiang2023unikgqa}
Jinhao Jiang, Kun Zhou, Wayne~Xin Zhao, and Ji-Rong Wen.
\newblock Unikgqa: Unified retrieval and reasoning for solving multi-hop question answering over knowledge graph.
\newblock \emph{ICLR 2023}, 2023{\natexlab{b}}.

\bibitem[Kirkpatrick et~al.(2017)Kirkpatrick, Pascanu, Rabinowitz, Veness, Desjardins, Rusu, Milan, Quan, Ramalho, Grabska-Barwinska, et~al.]{kirkpatrick2017overcoming}
James Kirkpatrick, Razvan Pascanu, Neil Rabinowitz, Joel Veness, Guillaume Desjardins, Andrei~A Rusu, Kieran Milan, John Quan, Tiago Ramalho, Agnieszka Grabska-Barwinska, et~al.
\newblock Overcoming catastrophic forgetting in neural networks.
\newblock \emph{Proceedings of the national academy of sciences}, 114\penalty0 (13):\penalty0 3521--3526, 2017.

\bibitem[Li and Hoiem(2017)]{li2017learning}
Zhizhong Li and Derek Hoiem.
\newblock Learning without forgetting.
\newblock \emph{IEEE transactions on pattern analysis and machine intelligence}, 40\penalty0 (12):\penalty0 2935--2947, 2017.

\bibitem[Liang and Li(2024)]{liang2024inflora}
Yan-Shuo Liang and Wu-Jun Li.
\newblock Inflora: Interference-free low-rank adaptation for continual learning.
\newblock In \emph{Proceedings of the IEEE/CVF Conference on Computer Vision and Pattern Recognition}, pages 23638--23647, 2024.

\bibitem[Lin et~al.(2019)Lin, Chen, Chen, and Ren]{lin-etal-2019-kagnet}
Bill~Yuchen Lin, Xinyue Chen, Jamin Chen, and Xiang Ren.
\newblock {K}ag{N}et: Knowledge-aware graph networks for commonsense reasoning.
\newblock In \emph{EMNLP-IJCNLP}, pages 2829--2839, 2019.

\bibitem[Liu et~al.(2022)Liu, Gu, Chi, Wang, Yu, Chen, and Tang]{liu2022few}
Huan Liu, Li Gu, Zhixiang Chi, Yang Wang, Yuanhao Yu, Jun Chen, and Jin Tang.
\newblock Few-shot class-incremental learning via entropy-regularized data-free replay.
\newblock In \emph{European Conference on Computer Vision}, pages 146--162. Springer, 2022.

\bibitem[Liu et~al.(2023)Liu, Cao, Lu, wen Xiao, Bagdanov, and Cheng]{liu2023clipCIL}
Xialei Liu, Xusheng Cao, Haori Lu, Jia wen Xiao, Andrew~D. Bagdanov, and Ming-Ming Cheng.
\newblock Class incremental learning with pre-trained vision-language models, 2023.

\bibitem[Liu et~al.(2019)Liu, Li, Garcia-Duran, Niepert, Onoro-Rubio, and Rosenblum]{liu2019mmkg}
Ye Liu, Hui Li, Alberto Garcia-Duran, Mathias Niepert, Daniel Onoro-Rubio, and David~S Rosenblum.
\newblock Mmkg: multi-modal knowledge graphs.
\newblock In \emph{The Semantic Web: 16th International Conference, ESWC 2019, Portoro{\v{z}}, Slovenia, June 2--6, 2019, Proceedings 16}, pages 459--474. Springer, 2019.

\bibitem[Mallya et~al.(2018)Mallya, Davis, and Lazebnik]{mallya2018piggyback}
Arun Mallya, Dillon Davis, and Svetlana Lazebnik.
\newblock Piggyback: Adapting a single network to multiple tasks by learning to mask weights.
\newblock In \emph{ECCV}, 2018.

\bibitem[Masana et~al.(2022)Masana, Liu, Twardowski, Menta, Bagdanov, and Van De~Weijer]{masana2022class}
Marc Masana, Xialei Liu, Bart{\l}omiej Twardowski, Mikel Menta, Andrew~D Bagdanov, and Joost Van De~Weijer.
\newblock Class-incremental learning: survey and performance evaluation on image classification.
\newblock \emph{IEEE Transactions on Pattern Analysis and Machine Intelligence}, 45\penalty0 (5):\penalty0 5513--5533, 2022.

\bibitem[McCloskey and Cohen(1989)]{mccloskey1989catastrophic}
Michael McCloskey and Neal~J Cohen.
\newblock Catastrophic interference in connectionist networks: The sequential learning problem.
\newblock In \emph{Psychology of learning and motivation}, pages 109--165. Elsevier, 1989.

\bibitem[Nayak and Bach(2022)]{nayaktmlr22}
N.~V. Nayak and S.~H. Bach.
\newblock Zero-shot learning with common sense knowledge graphs.
\newblock \emph{Transactions on Machine Learning Research (TMLR)}, 2022.

\bibitem[Pan et~al.(2024)Pan, Luo, Wang, Chen, Wang, and Wu]{pan2024unifying}
Shirui Pan, Linhao Luo, Yufei Wang, Chen Chen, Jiapu Wang, and Xindong Wu.
\newblock Unifying large language models and knowledge graphs: A roadmap.
\newblock \emph{IEEE Transactions on Knowledge and Data Engineering}, 2024.

\bibitem[Peng et~al.(2022)Peng, Zhao, Wang, Li, and Lovell]{peng2022few}
Can Peng, Kun Zhao, Tianren Wang, Meng Li, and Brian~C Lovell.
\newblock Few-shot class-incremental learning from an open-set perspective.
\newblock In \emph{European Conference on Computer Vision}, pages 382--397. Springer, 2022.

\bibitem[Petroni et~al.(2019)Petroni, Rockt{\"a}schel, Riedel, Lewis, Bakhtin, Wu, and Miller]{petroni2019language}
Fabio Petroni, Tim Rockt{\"a}schel, Sebastian Riedel, Patrick Lewis, Anton Bakhtin, Yuxiang Wu, and Alexander Miller.
\newblock Language models as knowledge bases?
\newblock In \emph{EMNLP-IJCNLP}, pages 2463--2473, 2019.

\bibitem[Radford et~al.(2021)Radford, Kim, Hallacy, Ramesh, Goh, Agarwal, Sastry, Askell, Mishkin, Clark, et~al.]{radford2021learning}
Alec Radford, Jong~Wook Kim, Chris Hallacy, Aditya Ramesh, Gabriel Goh, Sandhini Agarwal, Girish Sastry, Amanda Askell, Pamela Mishkin, Jack Clark, et~al.
\newblock Learning transferable visual models from natural language supervision.
\newblock In \emph{International conference on machine learning}, pages 8748--8763. PMLR, 2021.

\bibitem[Rebuffi et~al.(2017)Rebuffi, Kolesnikov, Sperl, and Lampert]{rebuffi2017icarl}
Sylvestre-Alvise Rebuffi, Alexander Kolesnikov, Georg Sperl, and Christoph~H Lampert.
\newblock icarl: Incremental classifier and representation learning.
\newblock In \emph{CVPR}, 2017.

\bibitem[Roy et~al.(2024)Roy, Moulick, Verma, Ghosh, and Das]{2024convprompt}
Anurag Roy, Riddhiman Moulick, Vinay Verma, Saptarshi Ghosh, and Abir Das.
\newblock Convolutional prompting meets language models for continual learning.
\newblock In \emph{Proceedings of the IEEE/CVF Conference on Computer Vision and Pattern Recognition (CVPR)}, 2024.

\bibitem[Serra et~al.(2018)Serra, Suris, Miron, and Karatzoglou]{serra2018overcoming}
Joan Serra, Didac Suris, Marius Miron, and Alexandros Karatzoglou.
\newblock Overcoming catastrophic forgetting with hard attention to the task.
\newblock In \emph{ICML}, 2018.

\bibitem[Shi et~al.(2021)Shi, Chen, Zhang, Zhan, and Wu]{shi2021overcoming}
Guangyuan Shi, Jiaxin Chen, Wenlong Zhang, Li-Ming Zhan, and Xiao-Ming Wu.
\newblock Overcoming catastrophic forgetting in incremental few-shot learning by finding flat minima.
\newblock \emph{Advances in neural information processing systems}, 34:\penalty0 6747--6761, 2021.

\bibitem[Smith et~al.(2023)Smith, Karlinsky, Gutta, Cascante-Bonilla, Kim, Arbelle, Panda, Feris, and Kira]{smith2023coda}
James~Seale Smith, Leonid Karlinsky, Vyshnavi Gutta, Paola Cascante-Bonilla, Donghyun Kim, Assaf Arbelle, Rameswar Panda, Rogerio Feris, and Zsolt Kira.
\newblock Coda-prompt: Continual decomposed attention-based prompting for rehearsal-free continual learning.
\newblock In \emph{Proceedings of the IEEE/CVF Conference on Computer Vision and Pattern Recognition}, pages 11909--11919, 2023.

\bibitem[Speer et~al.(2017)Speer, Chin, and Havasi]{speer2017conceptnet}
Robyn Speer, Joshua Chin, and Catherine Havasi.
\newblock Conceptnet 5.5: An open multilingual graph of general knowledge.
\newblock In \emph{Proceedings of the AAAI conference on artificial intelligence}, 2017.

\bibitem[Sun et~al.(2023{\natexlab{a}})Sun, Zhou, Ye, and Zhan]{sun2023pilot}
Hai-Long Sun, Da-Wei Zhou, Han-Jia Ye, and De-Chuan Zhan.
\newblock Pilot: A pre-trained model-based continual learning toolbox.
\newblock \emph{arXiv preprint arXiv:2309.07117}, 2023{\natexlab{a}}.

\bibitem[Sun et~al.(2023{\natexlab{b}})Sun, Xu, Tang, Wang, Lin, Gong, Shum, and Guo]{sun2023think}
Jiashuo Sun, Chengjin Xu, Lumingyuan Tang, Saizhuo Wang, Chen Lin, Yeyun Gong, Heung-Yeung Shum, and Jian Guo.
\newblock Think-on-graph: Deep and responsible reasoning of large language model with knowledge graph.
\newblock \emph{arXiv preprint arXiv:2307.07697}, 2023{\natexlab{b}}.

\bibitem[Tao et~al.(2020)Tao, Hong, Chang, Dong, Wei, and Gong]{tao2020few}
Xiaoyu Tao, Xiaopeng Hong, Xinyuan Chang, Songlin Dong, Xing Wei, and Yihong Gong.
\newblock Few-shot class-incremental learning.
\newblock In \emph{Proceedings of the IEEE/CVF Conference on Computer Vision and Pattern Recognition}, pages 12183--12192, 2020.

\bibitem[Thengane et~al.(2022)Thengane, Khan, Hayat, and Khan]{thengane2022continualclip}
Vishal Thengane, Salman Khan, Munawar Hayat, and Fahad Khan.
\newblock Clip model is an efficient continual learner.
\newblock \emph{arXiv:2210.03114}, 2022.

\bibitem[Vrande{\v{c}}i{\'c} and Kr{\"o}tzsch(2014)]{vrandevcic2014wikidata}
Denny Vrande{\v{c}}i{\'c} and Markus Kr{\"o}tzsch.
\newblock Wikidata: a free collaborative knowledgebase.
\newblock \emph{Communications of the ACM}, 57\penalty0 (10):\penalty0 78--85, 2014.

\bibitem[Wang et~al.(2022{\natexlab{a}})Wang, Zhou, Ye, and Zhan]{wang2022foster}
Fu-Yun Wang, Da-Wei Zhou, Han-Jia Ye, and De-Chuan Zhan.
\newblock Foster: Feature boosting and compression for class-incremental learning.
\newblock \emph{arXiv preprint arXiv:2204.04662}, 2022{\natexlab{a}}.

\bibitem[Wang et~al.(2023)Wang, Xie, Zhang, Huang, Su, and Zhu]{wang2023hide}
Liyuan Wang, Jingyi Xie, Xingxing Zhang, Mingyi Huang, Hang Su, and Jun Zhu.
\newblock Hierarchical decomposition of prompt-based continual learning: Rethinking obscured sub-optimality.
\newblock \emph{Advances in Neural Information Processing Systems}, 2023.

\bibitem[Wang et~al.(2022{\natexlab{b}})Wang, Zhang, Ebrahimi, Sun, Zhang, Lee, Ren, Su, Perot, Dy, et~al.]{wang2022dualprompt}
Zifeng Wang, Zizhao Zhang, Sayna Ebrahimi, Ruoxi Sun, Han Zhang, Chen-Yu Lee, Xiaoqi Ren, Guolong Su, Vincent Perot, Jennifer Dy, et~al.
\newblock Dualprompt: Complementary prompting for rehearsal-free continual learning.
\newblock In \emph{ECCV}, 2022{\natexlab{b}}.

\bibitem[Wang et~al.(2022{\natexlab{c}})Wang, Zhang, Lee, Zhang, Sun, Ren, Su, Perot, Dy, and Pfister]{l2p}
Zifeng Wang, Zizhao Zhang, Chen{-}Yu Lee, Han Zhang, Ruoxi Sun, Xiaoqi Ren, Guolong Su, Vincent Perot, Jennifer~G. Dy, and Tomas Pfister.
\newblock Learning to prompt for continual learning.
\newblock \emph{CVPR}, 2022{\natexlab{c}}.

\bibitem[Wu et~al.(2019)Wu, Chen, Wang, Ye, Liu, Guo, and Fu]{wu2019large}
Yue Wu, Yinpeng Chen, Lijuan Wang, Yuancheng Ye, Zicheng Liu, Yandong Guo, and Yun Fu.
\newblock Large scale incremental learning.
\newblock In \emph{Proceedings of the IEEE Conference on Computer Vision and Pattern Recognition}, pages 374--382, 2019.

\bibitem[Yan et~al.(2021)Yan, Xie, and He]{yan2021dynamically}
Shipeng Yan, Jiangwei Xie, and Xuming He.
\newblock Der: Dynamically expandable representation for class incremental learning.
\newblock In \emph{CVPR}, 2021.

\bibitem[Yu et~al.(2024)Yu, Zhuge, Zhang, Hu, Wang, Lu, and He]{yu2024boosting}
Jiazuo Yu, Yunzhi Zhuge, Lu Zhang, Ping Hu, Dong Wang, Huchuan Lu, and You He.
\newblock Boosting continual learning of vision-language models via mixture-of-experts adapters.
\newblock In \emph{Proceedings of the IEEE/CVF Conference on Computer Vision and Pattern Recognition}, pages 23219--23230, 2024.

\bibitem[Yu et~al.(2020)Yu, Twardowski, Liu, Herranz, Wang, Cheng, Jui, and Weijer]{yu2020semantic}
Lu Yu, Bartlomiej Twardowski, Xialei Liu, Luis Herranz, Kai Wang, Yongmei Cheng, Shangling Jui, and Joost van~de Weijer.
\newblock Semantic drift compensation for class-incremental learning.
\newblock In \emph{Proceedings of the IEEE/CVF conference on computer vision and pattern recognition}, pages 6982--6991, 2020.

\bibitem[Zhang et~al.(2021)Zhang, Song, Lin, Zheng, Pan, and Xu]{zhang2021few}
Chi Zhang, Nan Song, Guosheng Lin, Yun Zheng, Pan Pan, and Yinghui Xu.
\newblock Few-shot incremental learning with continually evolved classifiers.
\newblock In \emph{Proceedings of the IEEE/CVF conference on computer vision and pattern recognition}, pages 12455--12464, 2021.

\bibitem[Zhang et~al.(2023)Zhang, Wang, Kang, Chen, and Wei]{zhang2023slca}
Gengwei Zhang, Liyuan Wang, Guoliang Kang, Ling Chen, and Yunchao Wei.
\newblock Slca: Slow learner with classifier alignment for continual learning on a pre-trained model.
\newblock In \emph{Proceedings of the IEEE/CVF International Conference on Computer Vision}, 2023.

\bibitem[Zhang et~al.(2024)Zhang, Sun, Wang, Fan, Mo, Xu, Liu, Yang, and Shi]{zhang2024graphtranslator}
Mengmei Zhang, Mingwei Sun, Peng Wang, Shen Fan, Yanhu Mo, Xiaoxiao Xu, Hong Liu, Cheng Yang, and Chuan Shi.
\newblock Graphtranslator: Aligning graph model to large language model for open-ended tasks.
\newblock In \emph{Proceedings of the ACM on Web Conference 2024}, pages 1003--1014, 2024.

\bibitem[Zhang et~al.(2022)Zhang, Bosselut, Yasunaga, Ren, Liang, Manning, and Leskovec]{zhang2022greaselm}
Xikun Zhang, Antoine Bosselut, Michihiro Yasunaga, Hongyu Ren, Percy Liang, Christopher~D Manning, and Jure Leskovec.
\newblock Greaselm: Graph reasoning enhanced language models.
\newblock In \emph{ICLR}, 2022.

\bibitem[Zhao et~al.(2020)Zhao, Xiao, Gan, Zhang, and Xia]{zhao2020maintaining}
Bowen Zhao, Xi Xiao, Guojun Gan, Bin Zhang, and Shu-Tao Xia.
\newblock Maintaining discrimination and fairness in class incremental learning.
\newblock In \emph{CVPR}, 2020.

\bibitem[Zhou et~al.(2024)Zhou, Cai, Ye, Zhan, and Liu]{zhou2024revisiting}
Da-Wei Zhou, Zi-Wen Cai, Han-Jia Ye, De-Chuan Zhan, and Ziwei Liu.
\newblock Revisiting class-incremental learning with pre-trained models: Generalizability and adaptivity are all you need.
\newblock \emph{International Journal of Computer Vision}, 2024.

\bibitem[Zhou et~al.(2025)Zhou, Zhang, Wang, Ning, Ye, Zhan, and Liu]{zhou2023proof}
Da-Wei Zhou, Yuanhan Zhang, Yan Wang, Jingyi Ning, Han-Jia Ye, De-Chuan Zhan, and Ziwei Liu.
\newblock Learning without forgetting for vision-language models.
\newblock \emph{IEEE Transactions on Pattern Analysis and Machine Intelligence}, 2025.

\bibitem[Zhu et~al.(2023)Zhu, Chen, Shen, Li, and Elhoseiny]{zhu2023minigpt}
Deyao Zhu, Jun Chen, Xiaoqian Shen, Xiang Li, and Mohamed Elhoseiny.
\newblock Minigpt-4: Enhancing vision-language understanding with advanced large language models.
\newblock \emph{arXiv preprint arXiv:2304.10592}, 2023.

\bibitem[Zhu et~al.(2021)Zhu, Zhang, Wang, Yin, and Liu]{zhu2021prototype}
Fei Zhu, Xu-Yao Zhang, Chuang Wang, Fei Yin, and Cheng-Lin Liu.
\newblock Prototype augmentation and self-supervision for incremental learning.
\newblock In \emph{Proceedings of the IEEE/CVF Conference on Computer Vision and Pattern Recognition}, pages 5871--5880, 2021.

\end{thebibliography}
